\documentclass[runningheads]{llncs}
\usepackage{amsmath}
\usepackage{graphicx}
\usepackage{color}
\usepackage[table]{xcolor}
\usepackage{enumitem}
\usepackage{bm}
\usepackage{amsmath}
\usepackage{amssymb}
\usepackage{amsfonts}
\usepackage{bbding}
\definecolor{bblue}{rgb}{0,150,230}
\definecolor{mygray}{gray}{.91}
\definecolor{mygray2}{gray}{.95}
\definecolor{myy}{RGB}{126,95,0}

\makeatletter
\newcommand{\thickhline}{%
	\noalign {\ifnum 0=`}\fi \hrule height 1pt
	\futurelet \reserved@a \@xhline
}
\newcommand{\printfnsymbol}[1]{%
	\textsuperscript{\@fnsymbol{#1}}%
}
\makeatother

\makeatother
\newcommand{\figref}[1]{Fig.~\ref{#1}}

\newcommand{\eg}[1]{\textit{e.g.,}}
\newcommand{\ie}[1]{\textit{i.e.,}}

\usepackage[pagebackref=false,breaklinks=true,colorlinks=true,bookmarks=false,citecolor=blue]{hyperref}

\begin{document}
	
	\title{Quality-Aware Memory Network for \\ Interactive Volumetric Image Segmentation}
	
	%
	%
	\author{Tianfei Zhou\inst{1}\thanks{The first two authors contribute equally to this work.}, Liulei Li\inst{2}\printfnsymbol{1}, Gustav Bredell\inst{1}, Jianwu Li\inst{2}, Ender Konukoglu\inst{1}}
	
	\authorrunning{T. Zhou et al.}
	\titlerunning{Quality-Aware Memory Network for Interactive Image Segmentation}
	%
	%
	\institute{
		Computer Vision Laboratory, ETH Zurich, Switzerland  
		\email{\{tianfei.zhou,gustav.bredell,ender.konukoglu\}@vision.ee.ethz.ch} \and
		School of Computer Science and Technology, Beijing Institute of Technology, China \\
		\email{\{liliulei,ljw\}@bit.edu.cn} \\
	\url{https://github.com/0liliulei/Mem3D}	}
	\maketitle              

	\begin{abstract}
		
		Despite recent progress of automatic medical image segmentation techniques, fully automatic results usually fail to meet the clinical use and typically require further refinement. In this work, we propose a \textit{quality-aware memory network} for interactive segmentation of 3D medical images. Provided by user guidance on an arbitrary slice, an interaction network is firstly employed to obtain an initial 2D segmentation. The quality-aware memory network subsequently propagates the initial segmentation estimation bidirectionally over the entire volume. Subsequent refinement based on additional user guidance on other slices can be incorporated in the same manner. To further facilitate interactive segmentation, a quality assessment module is introduced to suggest the next slice to segment based on the current segmentation quality of each slice. The proposed network has two appealing characteristics: 1) The memory-augmented network offers the ability to quickly encode past segmentation information, which will be retrieved for the segmentation of other slices; 2) The quality assessment module enables the model to directly estimate the qualities of segmentation predictions, which allows an active learning paradigm where users preferentially label the lowest-quality slice for multi-round refinement. The proposed network leads to a robust interactive segmentation engine, which can generalize well to various types of user annotations (\eg, scribbles, boxes). Experimental results on various medical datasets demonstrate the superiority of our approach in comparison with existing techniques.
		
		\keywords{Interactive Segmentation \and Memory-Augmented Network}
	\end{abstract}
	\section{Introduction}
	Accurate segmentation of organs/lesions from medical imaging data holds the promise of significant improvement of clinical treatment, by allowing the extraction of accurate models for visualization, quantification or simulation. Although recent deep learning based automatic segmentation engines~\cite{ronneberger2015u,zhou2018unet++,wang2021exploring,milletari2016v} have achieved impressive performance, they still struggle to achieve sufficiently accurate and robust results for clinical practice, especially in the presence of poor image quality (\eg, noise, low contrast) or highly variable shapes (\eg, anatomical structures).
	Consequently, \textit{interactive segmentation}\!~\cite{olabarriaga2001interaction,zhao2013overview,zhou2017fixed,bredell2018iterative,wang2018deepigeos,wang2018interactive} garners research interests of the medical image analysis community, and recently became the choice in many real-life medical applications.


	In interactive segmentation, the user is factored in to play a crucial role in guiding the segmentation process and in correcting errors as they occur (often in an iteratively-refined manner). Classical approaches employ Graph Cuts\!~\cite{boykov2001interactive}, GeoS\!~\cite{criminisi2008geos} or Random Walker\!~\cite{grady2005random,grady2006random} to incorporate scribbles for segmentation. Yet, these methods require a large amount of input from users to segment targets with low contrast and ambiguous boundaries.
	With the advent of deep learning, there has been a dramatically increasing interest in learning from user interactions. Recent methods demonstrate higher segmentation accuracy with fewer user interactions than classical approaches. Despite this, many approaches\!~\cite{kitrungrotsakul2020interactive,sun2018interactive,sakinis2019interactive} only focus on 2D medical images, which are infeasible to process ubiquitous 3D data. Moreover, 2D segmentation does not allow the integration of prior knowledge regarding the 3D structure, and slice-by-slice interactive segmentation will impose extremely high annotation cost to users. To address this, many works~\cite{cciccek20163d,rajchl2016deepcut,liao2020iteratively,wang2018deepigeos,wang2018interactive} carefully design 3D networks to segment voxels at a time. While these methods enjoy superior ability of learning high-order, volumetric features, they require significantly more parameters and computations in comparison with the 2D counterparts. This necessitates
	compromises in the 3D network design to fit into a given memory or computation budget.
	


	
	To address these issues, we take a novel perspective to explore memory-augmented neural networks\!~\cite{sukhbaatar2015end,kumar2016ask,santoro2016meta,lu2020video} for 3D medical image segmentation. Memory networks augment neural networks with an external memory component, which allows the network to explicitly access the past experiences. They have been shown effective in few-shot learning\!~\cite{santoro2016meta}, contrastive learning\!~\cite{he2020momentum,wang2021exploring}, and also been explored to solve reasoning problems in visual dialog\!~\cite{sukhbaatar2015end,kumar2016ask}. The basic idea is to retrieve the relevant information from the external memory to answer a question at hand by using trainable memory modules. We take inspiration from these efforts to cast volumetric segmentation as a memory-based reasoning problem. Fundamental to our model is an external memory, which enables the model to online store segmented slices in the memory and 
	later mine useful representations from the memory for segmenting other slices. In this way, our model makes full use of context within 3D data, and at the same time, avoids computationally expensive 3D operations. During segmentation, we dynamically update the memory to maintain shape or appearance variations of the target. This facilitates easy model updating without extensive parameter optimization.
	Based on the memory network, we propose a novel interactive segmentation engine with three basic processes: 1)\!~\textit{Initialization:} an interaction network is employed to respond to user guidance on an arbitrary slice to obtain an initial 2D segmentation of a  target. 2)\!~\textit{Propagation:} the memory network propagates the initial mask to the entire volume. 3)\!~\textit{Refinement:} the physician could provide extra guidance on low-quality slices for iterative refinement if the segmentation results are unsatisfactory.

	%
	
	Our contributions are three-fold: \textbf{First}, we propose a memory-augmented network for volumetric interactive segmentation. It is able to incorporate rich 3D contextual information, while avoiding expensive 3D operations. \textbf{Second}, we equip the memory network  with a quality assessment module to assess the quality of each segmentation. This facilitates automatic selection of appropriate slices for iterative correction via human-in-the-loop. \textbf{Third}, our approach outperforms previous methods by a significant margin on two public datasets, while being able to handle various forms of interactions (\eg, scribbles, bounding boxes).

	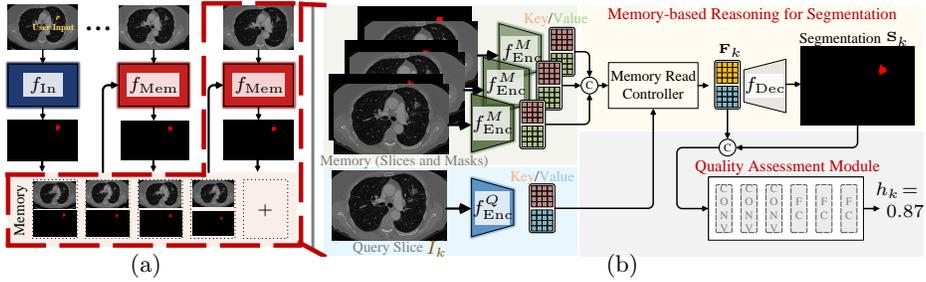
\begin{figure}[t]
		\begin{center}
			\includegraphics[width=\linewidth]{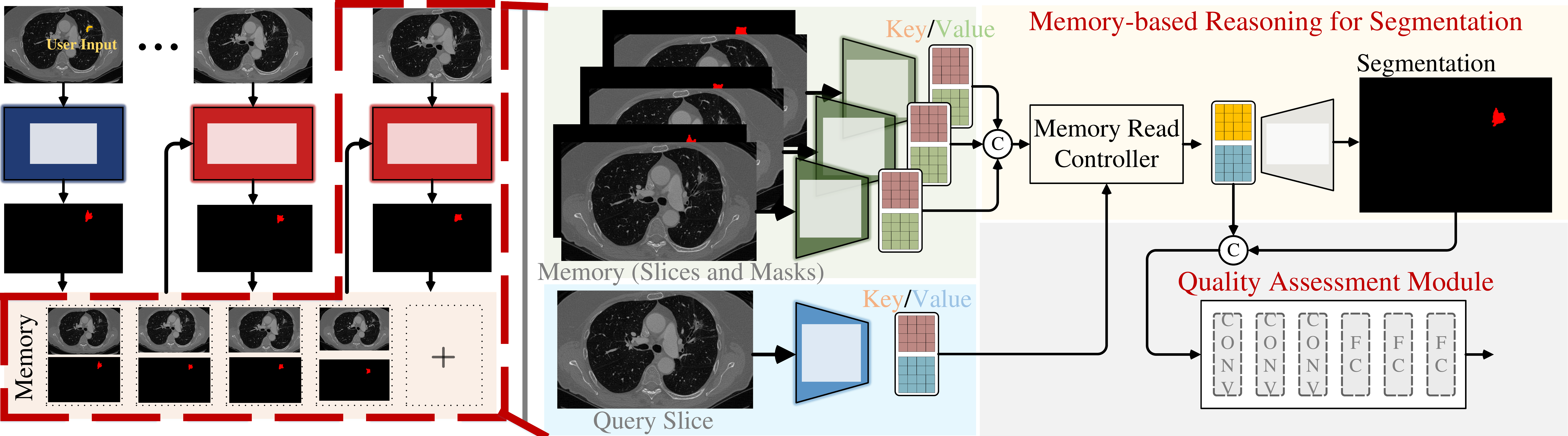}       
			\put(-338,63){\scriptsize $f_{\text{In}}$}		
			\put(-300.5,63){\scriptsize $f_{\text{Mem}}$}
			\put(-260.5,63){\scriptsize $f_{\text{Mem}}$}
			\put(-159.5,75.5){\scriptsize $f_{\text{Enc}}^M$}
			\put(-165.5,63){\scriptsize $f_{\text{Enc}}^M$}
			\put(-170.5,48.5){\scriptsize $f_{\text{Enc}}^M$}
			\put(-170.5,16.5){\scriptsize $f_{\text{Enc}}^Q$}
			\put(-67.5,63){\scriptsize $f_{\text{Dec}}$}
			\put(-14,81.5){\tiny $\mathbf{S}_k$}
			\put(-188,1){\scriptsize {\color{myy}{$I_k$}}}
			\put(-14,15){\scriptsize $0.87$}
			\put(-18,23){\scriptsize $h_k\!=$}
			\put(-300,-6){\small (a)}
			\put(-120,-6){\small (b)}
			\put(-77.5,77){\tiny $\mathbf{F}_k$}
		\end{center}
		\caption{Illustration of the proposed 3D interactive segmentation engine. (a) Simplified schematization of our engine that solves the task with an interaction network ($f_{\text{In}}$) and a quality-aware memory network ($f_{\text{Mem}}$). (b) Detailed network architecture of $f_{\text{Mem}}$. \copyright~denotes the concatenation operation. Zoom in for details.}
		\label{fig:framework}
	\end{figure}

	\section{Methodology}
	
	Let $V\in\mathbb{R}^{h\times w\times c}$ be a volumetric image to be segmented, which has a spatial size of $h\times w$ and $c$ slices. Our approach aims to obtain a 3D binary mask $\mathbf{S}\in\{0,1\}^{h\times w\times c}$ for a specified target by utilizing user guidance. As shown in~\figref{fig:framework}\!~(a), the physician is asked to provide an initial input on an arbitrary slice ${I}_i\in\mathbb{R}^{h\times w}$, where ${I}_i$ denotes the $i$-th slice of $V$. Then, an interaction network ($f_{\text{In}}$, \S\ref{sec:in}) is employed to obtain a coarse 2D segmentation $\mathbf{S}_i\in[0,1]^{h\times w}$ for ${I}_i$. Subsequently, $\mathbf{S}_i$ is propagated to all other slices with a quality-aware memory network ($f_{\text{Mem}}$, \S\ref{sec:memory}) to obtain $\mathbf{S}$. Our approach also takes into account iterative refinement so that segmentation performance can be progressively improved with multi-round inference. To aid the refinement, the memory network has a module that estimates the segmentation performance on each slice and suggests the user to place guidance on the slice with the worst segmentation quality. 
	
	
	%

	\subsection{Interaction Network} \label{sec:in}
	
	The interaction network takes the user annotation at an interactive slice ${I}_i$ to segment the specified target (or refine the previous result). At the $t^{th}$ round, its input consists of three images: the original gray-scale image ${I}_i$, the segmentation mask from the previous round $\mathbf{S}_i^{t-1}$, and a binary image $\mathbf{M}_i\!\in\!\{0,1\}^{h\times w}$ that encodes user guidance. Note that in the first round (\ie, $t\!=\!0$), the segmentation mask $\mathbf{S}_i^{-1}$ is initialized as a neutral mask with $0.5$ for all pixels. These inputs are concatenated along the channel dimension to form an input tensor $\mathbf{X}_i^{t}\!\in\!\mathbb{R}^{h\times w\times 3}$. The interaction network $f_{\text{IN}}$ conducts the segmentation for $I_i$ as follows:
	\begin{equation}\label{eq:in}
		\begin{aligned}
			\mathbf{S}_i^t = f_{\text{In}}(\mathbf{X}_i^t) \in \mathbb{R}^{h \times w}.
		\end{aligned}
	\end{equation}
	\subsubsection{Region-of-interest (ROI).} To further enhance performance and avoid mistakes in case of small targets or low-contrast tissues, we propose to crop the image according to the rough bounding-box estimation of user input, and apply $f_{\text{IN}}$ only to the ROI. We extend the bounding box by 10\% along sides to preserve more context. Each ROI region is resized into a fixed size for network input. After segmentation, the mask made within the ROI is inversely warped and pasted back to the original location.
	


	\subsection{Quality-Aware Memory Network}\label{sec:memory}
	
	Given the initial segmentation $\mathbf{S}_i^{t}$, our memory network learns from the interactive slice ${I}_i$ and segments  the specified target in other slices.  It stores previously segmented slices in an external memory, and takes advantage of the stored 3D image and corresponding segmentation to improve the segmentation of each 2D query image. The network architecture is shown in \figref{fig:framework}(b). In the following paragraphs, the superscript `$t$' is omitted for conciseness unless necessary.
	

	\subsubsection{Key-Value Embedding.} Given a query slice ${I}_k$, the network mines useful information from memory $\mathcal{M}$ for segmentation. Here, each memory cell $\mathcal{M}_j\in\mathcal{M}$ consists of a slice ${I}_{m_j}$ and its segmentation mask $\mathbf{S}_{m_j}$, where $m_j$ indicates the index of the slice in the original volume. As shown in~\figref{fig:framework}(b), we first encode the query ${I}_k$ as well as each memory cell  $\mathcal{M}_j \!=\! \{{I}_{m_j},\!~\mathbf{S}_{m_j}\}$ into pairs of \textit{key} and \textit{value} using dedicated encoders (\ie, query $f_{\text{Enc}}^Q$ and memory encoder $f_{\text{Enc}}^M$):
	\begin{align}
		\mathbf{K}_k^Q, \mathbf{V}_k^Q &= f_{\text{Enc}}^Q ({I}_k), \label{eq:qenc} \\
		\mathbf{K}_{m_j}^M, \mathbf{V}_{m_j}^M & = f_{\text{Enc}}^M ({I}_{m_j}, \mathbf{S}_{m_j}). \label{eq:menc}
	\end{align}
	Here, $\mathbf{K}_k^Q\!\in\!\mathbb{R}^{H\times W\times C/8}$ and $\mathbf{V}_k^Q\!\in\!\mathbb{R}^{H\times W\times C/2}$ indicate key and value embedding of the query $I_k$, respectively, whereas $\mathbf{K}_{v_j}^M$ and $\mathbf{V}_{v_j}^M$ correspond to the key and value of the memory cell $\mathcal{M}_j$.  $H$, $W$ and $C$ denote the height, width and channel dimension of the feature map from the backbone network, respectively. Note that for each memory cell, we apply Eq.\!~\eqref{eq:menc} to obtain pairs of key and value embedding. Subsequently, all memory embedding are stacked together to build a pair of 4D key and value features (\ie,$\mathbf{K}^M\!\in\!\mathbb{R}^{N\times H\times W\times {C}/{8}}$ and $\mathbf{V}^M\!\in\!\mathbb{R}^{N\times H\times W\times C/2}$), where $N=|\mathcal{M}|$ denotes memory size. 
	
	\subsubsection{Memory Reading.} The memory read controller retrieves relevant information from the memory based on the current query. Following the key-value retrieval mechanism in\!~\cite{kumar2016ask,sukhbaatar2015end}, we first compute the similarity between every 3D location $p\in\mathbb{R}^3$ in $\mathbf{K}^M$ with each spatial location $q\in\mathbb{R}^2$ in $\mathbf{K}_k^Q$ with dot product:
	\begin{equation}
		s_k(p,q) = \frac{\mathbf{K}^M(p)\cdot\mathbf{K}_k^Q(q)}{\|\mathbf{K}^M(p)\|\|\mathbf{K}_k^Q(q)\|} \in [-1, 1],
	\end{equation}
	where $\mathbf{K}^M(p)\!\in\!\mathbb{R}^{C/8}$ and  $\mathbf{K}_k^Q(q)\!\in\!\mathbb{R}^{C/8}$ denote the features at the $p^{th}$ and $q^{th}$ position of $\mathbf{K}^M$ and $\mathbf{K}_k^Q$, respectively. 
	Next, we compute the read weight $w_k$ by softmax normalization:
	\begin{equation}
		w_k(p,q) = \exp(s_k(p,q)) / \sum\nolimits_o\exp(s_k(o,q)) \in [0, 1].
	\end{equation}
	Here, $w_k(p,q)$ measures the  matching probability between $p$ and $q$. The memory summarization is then obtained using the weight to  combine the memory value:
	\begin{equation}\label{eq:read}
		\mathbf{H}_k(q) = \sum\nolimits_p w_k(p,q) \mathbf{V}^M(p) \in \mathbb{R}^{C/2}.
	\end{equation}
	Here, $\mathbf{V}^M(p)\!\in\!\mathbb{R}^{C/2}$ denotes the feature of the $p^{th}$ 3D position in $\mathbf{V}^M$. $\mathbf{H}_k(q)$ indicates the summarized representation of location $q$. 
	For all $H\times W$ locations in $\mathbf{K}_k^Q$, we independently apply Eq.~\eqref{eq:read} and obtain the feature map $\mathbf{H}_k\in\mathbb{R}^{H\times W\times C/2}$. To achieve a more comprehensive representation, the feature map is concatenated with query value $\mathbf{V}_k^Q$ to compute a final representation $\mathbf{F}_k = \texttt{cat}(\mathbf{H}_k, \mathbf{V}_k^Q)\in\mathbb{R}^{H\times W\times C}$.
	
	\subsubsection{Final Segmentation Readout.} 
	$\mathbf{F}_k$ is leveraged by a decoder network $f_{\text{Dec}}$ to predict the final segmentation probability map for the query slice ${I}_k$:
	\begin{equation}
		\mathbf{S}_k = f_{\text{Dec}} (\mathbf{F}_k) \in [0,1]^{h\times w}.
	\end{equation}


	
	
	\subsubsection{Quality Assessment Module.}
	While the memory network provides a compelling way to produce 3D segmentation, it does not support human-in-the-loop scenarios. To this end, we equip the memory network with a lightweight quality assessment head, which computes a quality score for each segmentation mask. In particular, we consider \textit{mean intersection-over-union (mIoU)} as the basic index for quality measurement. For each query ${I}_k$, we take the feature $\mathbf{F}_k$ and the corresponding segmentation $\mathbf{S}_k$ together to regress a mIoU score $h_k$:
	\begin{equation}\label{eq:s}
		h_k = f_{\text{QA}} (\mathbf{F}_k, \mathbf{S}_k) \in [0,1], 
	\end{equation}
	where $\mathbf{S}_k$ is firstly resized to a size of $H\times W$ and then concatenated with $\mathbf{F}_k$ for decoding. The slice with the lowest score is curated for next-round interaction.
	
	\subsection{Detailed Network Architecture}
	
	We follow~\cite{zhang2020interactive} to implement the interaction network $f_{\text{In}}(\cdot)$ as a coarse-to-fine segmentation network, however, other network architectures (\eg, U-Net\!~\cite{ronneberger2015u}) can also be used here instead. The network is trained using the cross-entropy loss. For the quality-aware memory network, we utilize ResNet-50~\cite{he2016deep} as the backbone network for both $f_{\text{Enc}}^Q$ (Eq.~\eqref{eq:qenc}) and $f_{\text{Enc}}^M$ (Eq.~\eqref{eq:menc}). The \texttt{res4} feature map of ResNet-50 is taken for computing the key and value embedding. For $f_{\text{Dec}}(\cdot)$, we first apply Atrous Spatial Pyramid Pooling module after the memory read operation to enlarge the receptive field. We use three parallel dilated convolution layers with dilation rates 2, 4 and 8. Then, the learned feature is decoded with a residual refinement module proposed in~\cite{qin2019basnet}. The quality-aware module, $f_{\text{QA}}(\cdot)$, consists of three $3\!\times\!3$ convolutional layers and three fully connected layers.
	
	

	\section{Experiment}
	\subsubsection{Experimental Setup.}\label{sec:setup}
	Our experiments are conducted on two public datasets:
	\textbf{\textit{MSD}}\!~\cite{simpson2019large} includes ten subsets with different anatomy of interests, with a total of 2,633 3D volumes. In our experiments,  we study the most challenging lung ($64/32$ for \texttt{train}/\texttt{val}) and colon ($126/64$ for \texttt{train}/\texttt{val}) subsets.  \textbf{\textit{KiTS$_{19}$}}\!~\cite{heller2019kits19} contains $300$ arterial phase abdominal CT scans with annotations of kidney and tumor. We use the released $210$ scans ($168/42$ for \texttt{train}/\texttt{val}) for experiments.
	
	
	
	For comparison, we build a baseline model, named {{Interactive 3D nnU-Net}}, by adapting nnU-Net\!~\cite{isensee2018nnu} into an interactive version. Specifically, we use the interaction network (\S\ref{sec:in}) to obtain an initial segmentation, and this segment is then concatenated with the volume as the input of 3D nnU-Net. The quality-aware iterative refinement is also applied. In addition, we compare with a state-of-the-art method DeepIGeoS~\cite{wang2018deepigeos}. Several non-interactive methods are also included.
	
	\subsubsection{Interaction Simulation.} 
	Our approach can support various types of user interactions, which facilitates use in clinical routine. We study three common interactions: \textit{\textbf{Scribbles}} provide sparse labels to describe the targets and rough outreach, \textit{\textbf{Bounding Boxes}} outline the sizes and locations of targets, whereas \textit{\textbf{Extreme Points}}\!~\cite{maninis2018deep} outline a more compact area of a target by labeling its \textit{leftmost}, \textit{rightmost}, \textit{top}, \textit{bottom} pixels. To simulate scribbles, we manually label the data in KiTS$_{19}$ and MSD, resulting in $3,\!585$ slices. Bounding boxes and extreme points can be easily simulated from ground-truths with relaxations. We train an independent $f_{\text{In}}$ for each of these interaction types. In the first interaction round, we compute a rough ROI according to user input. Then we treat all the pixels out of the enlarged ROI as the background, and the pixels specified by scribbles, or in regions of bounding boxes and extreme points as foreground. We encode user guidance as a binary image $\mathbf{M}$ (\S\ref{sec:in}) for input.
	
		\begin{table}[t]
		\centering
		\caption{Quantitative results (DSC \%) on (left) MSD\!~\cite{simpson2019large} and (right) KiTS$_{19}$\!~\cite{heller2019kits19} \texttt{val}. }
		
		\resizebox{0.45\textwidth}{!}{%
			\begin{tabular}{l|ccc}

				method & lung cancer & colon cancer \\ 
				\thickhline
				\multicolumn{3}{l}{{\textit{non-interactive methods:}}} \\
				
				\hline
				
				C2FNAS\!~\cite{yu2020c2fnas} & 70.4 & 58.9 \\
				3D nnU-Net\!~\cite{isensee2018nnu} & 66.9 & 56.0 \\
				\hline
				
				\multicolumn{3}{l}{\textit{interactive methods:}} \\ 
				\hline
				\multicolumn{3}{c}{{Interactive 3D nnU-Net~\cite{isensee2018nnu}}} \\ \hline
				scribbles &  73.9  &  68.1  \\
				bounding boxes &  74.7  &  68.5  \\
				extreme points  & 75.1 & 69.8 \\ \hline
				
				\multicolumn{3}{c}{{DeepIGeoS~\cite{wang2018deepigeos}}} \\ \hline
				scribbles &  76.6  &  72.3  \\
				bounding boxes &  77.2  &  73.0  \\
				extreme points  & 77.5 & 73.2 \\ \hline

				\multicolumn{3}{c}{{\textbf{Ours}}} \\ \hline
				scribbles &  80.9  &  79.7  \\
				bounding boxes &  81.5  &  79.3  \\
				extreme points  & \textbf{82.0} & \textbf{80.4} \\

		\end{tabular}} 
		\resizebox{0.51\textwidth}{!}{%
			\begin{tabular}{l|ccc}
				method & kidney\!~(organ) & kidney\!~(tumor) \\ 
				\thickhline
				\multicolumn{3}{l}{\textit{non-interactive methods}} \\ \hline
				
				Mu et al.\!~\cite{mu2019segmentation} & \textbf{97.4} &  78.9 \\ 
				3D nnU-Net\!~\cite{isensee2018nnu} & 96.9 & 85.7 \\
				\hline
				
				\multicolumn{3}{l}{\textit{interactive methods}} \\ 
				\hline
				\multicolumn{3}{c}{{Interactive 3D nnU-Net~\cite{isensee2018nnu}}} \\ \hline
				scribbles &  94.5  &  86.3  \\
				bounding boxes &  95.3  &  86.8  \\
				extreme points & 95.6 & 87.6 \\ \hline
				
				\multicolumn{3}{c}{{DeepIGeoS~\cite{wang2018deepigeos}}} \\ \hline
				scribbles &  95.7  &  87.6  \\
				bounding boxes &  96.4  &  88.5  \\
				extreme points  & 96.7 & 88.9 \\ \hline

				\multicolumn{3}{c}{{\textbf{Ours}}} \\ \hline
				scribbles &  96.9  &  88.2  \\
				bounding boxes &  97.0 &  88.4  \\
				extreme points & 97.0 & \textbf{89.1} \\ 
				
		\end{tabular}} 
		\label{table:1}
	\end{table}
	
	\begin{figure}[t]
		\begin{center}
			\includegraphics[width=0.96\linewidth]{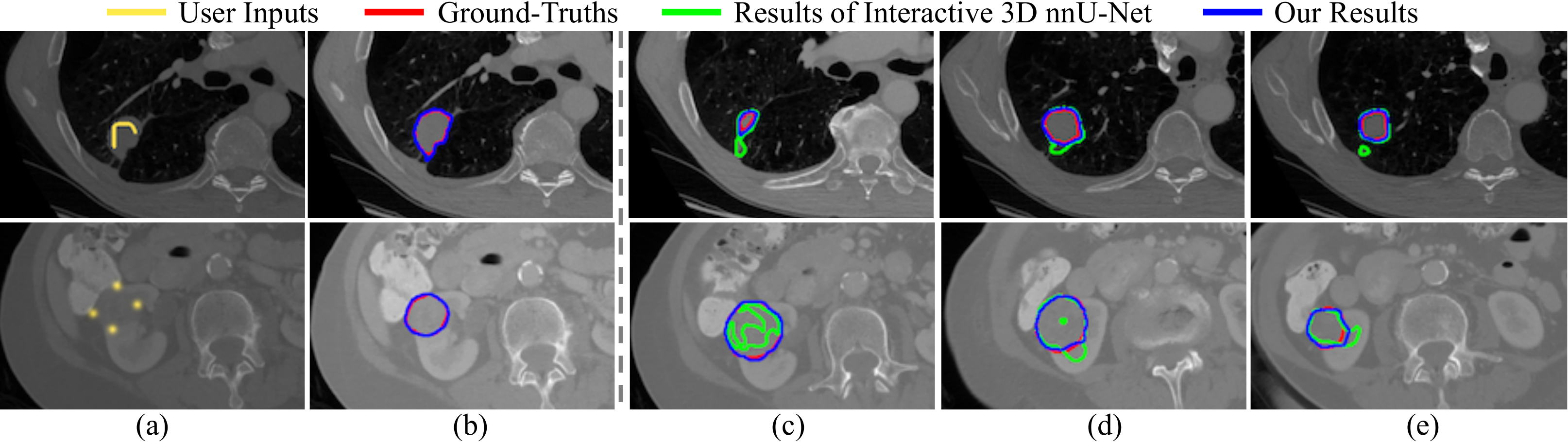}
		\end{center}
		\caption{Qualitative results of our approach \textit{v.s.} Interactive 3D nnU-Net on two samples in MSD-Lung (row \#1) and KITS$_{19}$ (row \#2), using scribbles and extreme points as supervision, respectively. (a) Interactive slices; (b) Results of interactive slices using the interaction network; (c)-(e) Results of other slices. Zoom in for details.}
		\label{fig:visual}
	\end{figure}
	
	\subsubsection{Training and Testing Details.}Our engine is implemented in PyTorch. We use the same settings as\!~\cite{zhang2020interactive} to train $f_{\text{In}}$\!~(\S\ref{sec:in}). The quality-aware memory network $f_{\text{Mem}}$ (\S\ref{sec:memory}) is trained using Adam with learning rate 1e-5 and batch size $8$ for $120$ epochs. To make a training sample, we randomly sample $5$ temporally ordered slices from a 3D image. During training, the memory is dynamically updated by adding the slice and mask at the previous step to the memory for the next slice.
	
	During inference, simulated user hints are provided to $f_{\text{IN}}$ for an initial segmentation of the interactive slice. Then, for each query slice, we put this interactive slice and the previous slice with corresponding segmentation mask into the memory as the most important reference information. In addition, we save a new memory item every $N$ slices, where $N$ is empirically set to $5$. We do not add all slices and corresponding masks into memory to avoid large storage and computational costs. In this way, our memory network achieves the effect of online learning and adaption without additional training.

	\subsubsection{Quantitative and Qualitative Results.}
	Table~\ref{table:1} (left) reports segmentation results of various methods on MSD \texttt{val}. For interactive methods, we report results at the $6^{th}$ round which well balances accuracy and efficiency. It can be seen that our method leads to consistent performance gains over the baselines. Specifically, our approach significantly outperforms Interactive 3D nnU-Net by more than \textbf{7\%} for lung cancer and \textbf{10\%} for colon cancer, and outperforms DeepIGeoS~\cite{wang2018deepigeos} by more than \textbf{4\%} and \textbf{7\%}, respectively. Moreover, for different types of interaction, our method produces very similar performance, revealing its high robustness to user input. Table~\ref{table:1} (right) presents performance comparisons on KiTS$_{19}$ \texttt{val}. The results demonstrate that, for kidney tumor segmentation, our engine generally outperforms the baseline models. The improvements are lower than seen for the MSD dataset due to the fact that the initial segmentation is already of high quality resulting in smaller dice score gains for adjustments.
	
	
	\figref{fig:visual} depicts qualitative comparisons of our approach against Interactive 3D nnU-Net on representative examples from MSD and KITS$_{19}$. As seen, our approach produces more accurate segmentation results than the competitor.

	\begin{table}[t]
		\centering
		
		\caption{Ablation study of the quality assessment module in terms of DSC (\%).}
		\resizebox{0.77\textwidth}{!}{%
			\setlength\tabcolsep{2pt}
			\renewcommand\arraystretch{1.0}
			\begin{tabular}{c|ccc}
				
				variant & MSD (lung) &  MSD (colon) & KiTS$_{19}$ (tumor) \\ 
				\thickhline
				oracle & 81.4 & 80.4 & 89.1 \\
				random & 80.1 & 77.5 & 86.8 \\ \hline
				\textbf{quality assess.}& 81.3 & 79.7 & 88.6 \\

		\end{tabular}} 
		\label{table:2}
	\end{table}

	\begin{figure}[t]
		\begin{center}
			\includegraphics[width=0.77\linewidth]{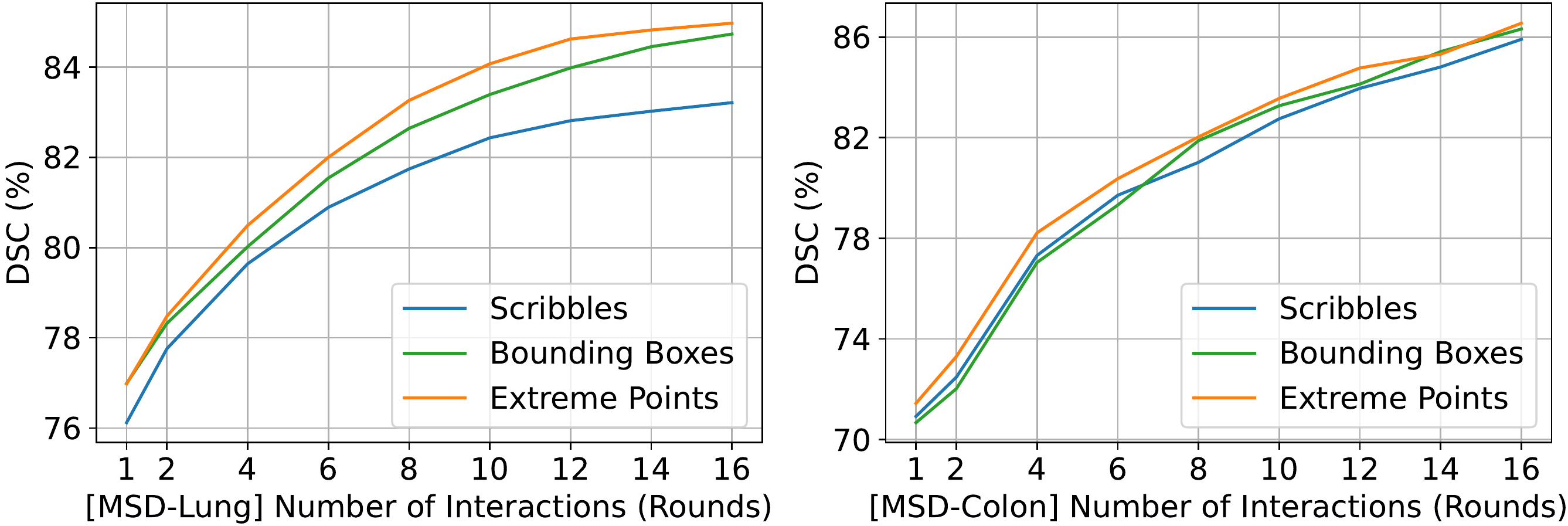}
		\end{center}
		\caption{The impact of number of interactions on MSD Lung (left) and Colon (right).}
		\label{fig:round}
	\end{figure}

	\subsubsection{Efficacy of Quality Assessment Module.}
	The quality assessment module empowers the engine to automatically select informative slices for iterative correction. To prove its efficacy, we design two baselines: `oracle' selects the worst segmented slice by comparing the masks with corresponding ground-truths, while `random' conducts random selection. As reported in Table~\ref{table:2}, our method (\ie, quality assessment module) significantly outperforms `random' across three sets, and is comparable to `oracle', proving its effectiveness.
	
	\subsubsection{Impact of Multi-Round Refinement.} 
	\figref{fig:round} shows DSC scores with growing number of interactions on lung and colon subsets of MSD. We observe that multi-round refinement is crucial for achieving higher segmentation performance, and the performance becomes almost marginal at the $16^{th}$ round. 
	
	\subsubsection{Runtime Analysis.} 
	For a 3D volume with size $512\!\times\!512\!\times\!100$, our method needs 5.13 s on average for one-round segmentation on a NVIDIA 2080Ti GPU, whereas Interactive 3D nnU-Net needs 200 s. Hence our engine enables a significant increase in inference speed.
	
	

	\section{Conclusion}
	
	This work presents a novel interactive segmentation engine for 3D medical volumes. The key component is a memory-augmented neural network, which employs an external memory for accurate and efficient 3D segmentation. Moreover, the quality-aware module empowers the engine to automatically select informative slices for user feedback, which we believe is an important added value of the memory network. Experiments on two public datasets show that our engine outperforms other alternatives while having a much faster inference speed.

	
	\bibliographystyle{splncs04}
	\bibliography{mybibliography}
	
\end{document}